\documentclass[conference]{IEEEtran}
\IEEEoverridecommandlockouts

\usepackage{cite}
\usepackage{url}
\usepackage[hidelinks]{hyperref}
\usepackage{amsmath,amssymb,amsfonts}
\usepackage{graphicx}
\usepackage{textcomp}
\usepackage[table]{xcolor}
\usepackage{booktabs}
\usepackage{orcidlink}
\usepackage{multirow}
\usepackage{makecell}
\usepackage{txfonts}    
\usepackage{tikz}
\usetikzlibrary{matrix,positioning,fit,backgrounds,calc}

\definecolor{cArchChan}  {rgb}{0.45,0.65,0.90}   
\definecolor{cArchScal}  {rgb}{1.00,0.82,0.10}   
\definecolor{cArchSeq}   {rgb}{0.20,0.72,0.33}   
\definecolor{cArchSet}   {rgb}{0.95,0.58,0.15}   
\definecolor{cArchTrunk} {rgb}{0.25,0.78,0.78}   
\definecolor{cArchPol}   {rgb}{0.72,0.45,0.90}   
\definecolor{cArchVal}   {rgb}{1.00,0.50,0.50}   

\colorlet{cHand}{cArchChan}
\colorlet{cOpen}{cArchSet}
\colorlet{cOwnDisc}{cArchTrunk}
\colorlet{cOppDisc}{cArchPol}
\colorlet{cOppPick}{cArchVal}
\colorlet{cWild}{cArchScal}
\definecolor{cSeen}     {rgb}{0.60,0.60,0.60}    
\definecolor{cLive}     {rgb}{0.72,0.94,0.25}    
\definecolor{cPartial}  {rgb}{1.00,0.55,0.42}    

\colorlet{cPureSeq}{cArchSeq}
\definecolor{cImpureSeq}{rgb}{0.35,0.25,0.85}    
\definecolor{cPureSet}  {rgb}{0.85,0.20,0.45}    
\colorlet{cDeadwood}{cSeen}
\colorlet{cMeld}{cArchChan}    

\newcommand{\hs}{{\color{red}\scalebox{1.2}{$\varheartsuit$}\hspace{0.05em}}}
\newcommand{\ds}{{\color{red}\scalebox{1.2}{$\vardiamondsuit$}\hspace{0.05em}}}
\newcommand{\cs}{{\scalebox{1.2}{$\clubsuit$}\hspace{0.05em}}}
\newcommand{\sps}{{\scalebox{1.2}{$\spadesuit$}\hspace{0.05em}}}

\begin{document}
\pagestyle{empty}

\title{IRumAI: Reinforcement Learning for Indian Rummy}

\author{\IEEEauthorblockN{Vignesh Mohan~\orcidlink{0000-0002-1295-7286}\thanks{To appear in Proceedings of the IEEE Conference on Games (CoG), Madrid, Spain, September 2026.}\thanks{\copyright2026 IEEE. Personal use of this material is permitted. Permission from IEEE must be obtained for all other uses, including reprinting/republishing this material for advertising or promotional purposes, creating new collective works for resale or redistribution to servers or lists, or reuse of any copyrighted component of this work in other works.}}
\IEEEauthorblockA{\textit{EURECOM}\\
Sophia Antipolis, France\\
vignesh.mohan@eurecom.fr}}

\IEEEpubid{\makebox[\columnwidth]{979-8-3315-9476-3/26/\$31.00 \copyright2026 IEEE\hfill}
\hspace{\columnsep}\makebox[\columnwidth]{ }}

\maketitle
\thispagestyle{empty}

\IEEEpubidadjcol

\begin{abstract}
  Despite its massive player base and complex hidden-information dynamics, Indian Rummy has received no reinforcement learning attention. Existing agents rely on combinatorial search, which is tactically strong but slow at inference. We present IRumAI, the first RL agent for the domain. IRumAI integrates Proximal Policy Optimization (PPO), meld-aware observation encoding, deadwood-driven reward shaping, and a dual-branch convolutional architecture. IRumAI is RL-trained solely against weak heuristics, after a one-time behaviour-cloning warm-start on stronger demonstration data. It generalises to defeat the entire baseline hierarchy, including a 53.9\% win rate against the strongest search-based opponent unseen during RL training. Bypassing explicit search, IRumAI requires just 0.33\,ms per action, which is over 7,000$\times$ faster than the state-of-the-art heuristic. Ablations validate our architectural choices, and linear probing reveals that the network implicitly models the opponent's hidden hand from public interactions.
\end{abstract}


\section{Introduction}

Card games have long served as important benchmarks for artificial intelligence research, providing a compact and rule-governed setting for studying hidden information, stochastic transitions, and long-horizon sequential decision-making~\cite{moravcik2017deepstack}. Among such games, Indian Rummy stands out for both its widespread popularity and its computational depth. Despite being widely played across South Asia with hundreds of millions of players~\cite{saha2025rummy}, it has remained outside the focus of the reinforcement learning community.

The central difficulty of Indian Rummy, like all its variants, is its strict meld-validity constraint. Unlike poker, where any drawn hand has a well-defined rank on a continuous spectrum of strength, an Indian Rummy hand is functionally worthless until the player can partition all thirteen of their cards into valid sequences and sets. The joint deal space alone exceeds $7\times10^{22}$ distinct configurations (see §\ref{sec:formulation}), and a randomly designated wildcard that rewrites the value of individual cards every game makes the per-game state distribution highly variable. Navigating this environment requires simultaneously building toward a valid combinatorial partition, inferring which cards the opponent is collecting from the open discard pile, and deciding when to block the opponent versus when to accelerate one's own progress.

The computational treatment of rummy-family games remains sparse compared to classic benchmarks like chess, Go, or poker. Most existing literature focuses on gin rummy, a two-player American variant where players end the game by knocking once their unmelded cards fall below a certain point threshold. Recent work has produced several rule-based and evaluation-function agents for this game. For instance, Goldman et~al.~\cite{goldman2021mmd} combine a myopic meld-distance heuristic with a rule-based opponent model, while other studies have explored discard-evaluation functions~\cite{eicholtz2021heisenbot,gallucci2021fitness} and deterministic neural networks for draw, discard, and knock decisions~\cite{nguyen2021ginnn}. Beyond gin rummy, Eswaran~et~al.~\cite{eswaran2020gaim} applied convolutional neural networks to classify player strategies in online rummy, demonstrating the viability of learned card representations. Their work, however, targets supervised pattern mining rather than policy learning. To our knowledge, no prior work applies iterative reinforcement learning to any rummy variant.

Classic Indian Rummy has received less attention. The prior study by Saha et~al.~\cite{saha2025rummy} formalised two core hand-evaluation metrics to drive rule-based play. The first metric, \textit{MinScore}, calculates the minimum deadwood score achievable by optimally partitioning the current hand. While it evaluates immediate hand value, it often plateaus because many incomplete hands result in the same maximum penalty. To address this, they introduced \textit{MinDist}, which measures the combinatorial edit distance required to transform the current hand into a valid winning declaration. Using these metrics, they established a hierarchy of non-learning, heuristic agents that serve as the baseline skill ladder for our experiments, ranging from a uniform random baseline to the strongest agent \textit{MinDistOppAgent}, which combines distance optimisation with heuristic tracking of the opponent's draws and discards.

Beyond rummy, the broader landscape of card-game AI offers the techniques we build on. Poker has been the canonical benchmark for imperfect-information games, where systems like DeepStack~\cite{moravcik2017deepstack} achieved superhuman play by pairing Counterfactual Regret Minimisation with online re-solving. These methods are theoretically principled for two-player zero-sum games, but their computational cost scales poorly with larger branching factors. This has motivated a shift toward model-free reinforcement learning, demonstrated by DouZero~\cite{zha2021douzero} (DouDizhu) and Multi-DMC~\cite{multidmc2025} (UNO), which mastered their games through deep Monte Carlo methods with highly concurrent rollouts. From this lineage we adopt two techniques: imitation learning to warm-start the policy in our large combinatorial space, following AlphaStar~\cite{vinyals2019alphastar}, and potential-based reward shaping~\cite{ng1999shaping} to address the sparse terminal rewards typical of card games while theoretically preserving the optimal policy.

In this work, we close this gap by introducing the first reinforcement learning agent for 13-card two-player Indian Rummy. IRumAI (\textbf{I}ndian \textbf{Rum}my \textbf{AI}) learns via PPO self-play against a fixed pool of weak but fast heuristic baselines, after a one-time behaviour-cloning warm-start, and operates without any hand-crafted search or lookahead at inference time. To support this training regime, we developed a PettingZoo~\cite{terry2021pettingzoo} environment that provides the policy with a rich ten-channel observation encoding. This encoding explicitly surfaces meld coverage, live outs, and partial-meld progress. By embedding these structural priors directly into the observation space, we obviate explicit tree search at inference, yielding sub-millisecond decision latencies.

We pair this with a PPO~\cite{schulman2017ppo} training pipeline driven by meld-aware, potential-based reward shaping~\cite{ng1999shaping} that densifies the sparse terminal rewards without altering the optimal policy. After 10{,}000 training updates, our agent outperforms the \textit{MinScore} and \textit{MinDist} family of heuristic baselines, including a 53.9\% win rate against the opponent-modelling heuristic \textit{MinDistOppAgent}, which it never encounters during training. Because the agent replaces combinatorial search with a single neural network forward pass, it requires only 0.33 milliseconds per action on a single ARM Neoverse-V2 core. This is over 7{,}000 times faster than \textit{MinDistOppAgent}. We also conduct ablation studies on reward shaping, behaviour-cloning initialisation, opponent-curriculum design, and policy architecture. Linear probing further reveals that the policy trunk encodes implicit opponent-hand information. We release our environment and training codebase to support further research.

\section{Background}

\subsection{Card Game: Indian Rummy}
\label{sec:game}

Indian Rummy is a draw-and-discard card game played widely across South Asia. The 13-card variant is played with a standard 52-card deck consisting of four suits (\hs~Hearts, \ds~Diamonds, \cs~Clubs, \sps~Spades), each containing 13 ranks (Ace, 2--10, Jack, Queen, King). In the two-player version (which is the focus of our setting), a single deck is used, and each player is dealt 13 cards face-down at the start of the episode. The remaining cards form the \emph{closed deck}; the top card of the closed deck is turned face-up to seed the \emph{open discard pile}.

\subsubsection{The Wildcard Joker}
Before play begins, a single card is drawn at random from the closed deck and placed aside, and its rank designates the \emph{wildcard} for that game.\footnote{Most commercial rummy variants include two printed jokers in the deck. Our simulation omits them, as printed jokers are structurally equivalent to a pre-designated wildcard rank and add minimal strategic complexity.} Every card matching this rank, regardless of suit, may substitute for any missing card in a meld. For example, if 9\cs\ is designated as the wildcard joker, then the 9\hs, 9\ds, 9\cs, and 9\sps\ all function as wildcards for that episode. Because this designation is drawn each game, the agent is prevented from exploiting a fixed card hierarchy.

\subsubsection{Turn Structure}
Players alternate turns, with each turn partitioned into two mandatory phases:
\begin{itemize}
  \item \textbf{Draw phase:} The player must take exactly one card, choosing either the face-up top card of the open pile or the unknown top card of the closed deck. Drawing from the open pile is visible to the opponent, whereas drawing from the closed deck is private.
  \item \textbf{Discard phase:} The player must place exactly one card from their hand face-up onto the open pile, returning their hand to exactly 13 cards. Alternatively, if the hand satisfies the declaration constraints (described below), the player may \emph{declare} to end the game.
\end{itemize}
The turn cycle continues until a player declares correctly, a player declares invalidly, or the closed deck is exhausted (resulting in a draw).

\subsubsection{Melds}
As in most rummy variants, the core objective is to form \emph{melds}, i.e., valid groupings of cards. Indian Rummy features four meld types:
\begin{itemize}
  \item \textbf{Pure sequence:} Three or more consecutive ranks of the same suit, using no wildcards. (\textit{Example}: \{3\hs, 4\hs, 5\hs, 6\hs\})
  \item \textbf{Impure sequence:} A consecutive run of the same suit in which one or more cards are replaced by wildcards. (\textit{Example, if wildcard=9}: \{3\hs, 4\hs, 9\cs, 6\hs\}, where 9\cs\ substitutes for the 5\hs.)
  \item \textbf{Pure set:} Three or four cards of the same rank in different suits, using no wildcards. (\textit{Example}: \{K\hs, K\ds, K\cs\})
  \item \textbf{Impure set:} A valid set in which one card is replaced by a wildcard. (\textit{Example, if wildcard=9}: \{K\hs, K\ds, 9\sps\}.)
\end{itemize}
Every meld must contain at least one natural (non-wildcard) card. Furthermore, a wildcard card cannot simultaneously serve in multiple melds.

\subsubsection{Declaration and Validity}
To win, a player must partition all 13 cards in their hand into valid melds and declare. A declaration is valid if and only if it satisfies two structural constraints:
\begin{enumerate}
  \item At least one meld is a \emph{pure sequence}.
  \item At least two melds are sequences (either pure or impure).
\end{enumerate}
Partial partitions are not accepted. All 13 cards must be covered. Attempting to declare when the hand does not satisfy these constraints constitutes an invalid declaration, resulting in an immediate loss regardless of the opponent's score.

\subsubsection{Deadwood and Scoring}

Cards not covered by any valid meld contribute \emph{deadwood points}. Numbered cards score their face value (2--9). Aces, tens, and face cards (Jack, Queen, King) each score 10 points. The maximum possible deadwood is 130 points (13 unmelded 10-point cards).

In practice, minimising deadwood serves as a dual proxy for minimising the penalty of losing and maximising the probability of a valid declaration. Every decision in Indian Rummy therefore requires balancing three competing objectives: (1)~reducing one's own deadwood toward zero, (2)~securing and maintaining the compulsory pure sequence, and (3)~denying the opponent useful cards via careful discarding. Drawing from the open pile accelerates one's hand, but telegraphs strategic information to the opponent. Holding a wildcard shortens the path to a valid declaration.

\section{Game Formulation}
\label{sec:formulation}

\subsection{Environment}

We formalise Indian Rummy as a two-player, zero-sum Markov game, implemented via the PettingZoo~\cite{terry2021pettingzoo} Agent Environment Cycle (AEC) API to handle the alternating turn structure. Each turn comprises two sequential decision nodes: a draw phase and a discard phase. The discrete action space contains 55 total actions, consisting of drawing from the open pile, drawing from the closed deck, discarding a specific card, and declaring. To prevent the agent from issuing illegal actions for the current phase, we enforce phase-constrained action masking. Only draw actions are valid during the draw phase, while only discard and declare actions are valid during the discard phase. The declare action is unmasked only when the agent's hand satisfies all validity constraints, namely at least one pure sequence, at least two sequences total, and a perfect 13-card partition. Because declaring is only available when the hand already constitutes a winning partition, the agent need only learn \emph{when} to declare rather than \emph{how}. Since a valid hand is always worth declaring immediately, this decision is trivial. The BC warm-start (§\ref{sec:opponents}) provides early exposure to this terminal transition.

The scale of the game can be quantified from a combinatorial lower bound on the initial deal. Each game begins by dealing 13 cards to each player from the 52-card deck. Because the two hands are drawn without replacement, the number of distinct joint hand configurations is $\binom{52}{13}\times\binom{39}{13} \approx 5.2\times10^{21}$. Since the wildcard rank is determined fresh each game by a separate draw from the remaining cards, this figure is further multiplied by up to 13, giving a joint deal space of roughly $\mathbf{7\times10^{22}}$ at game start alone. The full game-state space, which must additionally account for the sequence of discards, the ordering of the closed deck, and the current turn, is orders of magnitude larger, making exhaustive search infeasible.

\subsection{Observation and Hidden Information}

%


\begin{figure*}[t]
  \centering
  \resizebox{\linewidth}{!}{%
    \begin{tikzpicture}[
        font=\sffamily\scriptsize,
        >=stealth,
        x=0.45cm, y=0.45cm
      ]

      \newcommand{\pc}[4]{%
        \fill[#4, rounded corners=1.5pt]
        ({#1*13.5 + #2 + 0.15}, {-#3 - 0.15}) rectangle
        ({#1*13.5 + #2 + 0.85}, {-#3 - 0.85});
      }

      \foreach \s/\sym/\col in {0/\varheartsuit/red, 1/\vardiamondsuit/red, 2/\clubsuit/black, 3/\spadesuit/black} {
        \def\xoff{\s * 13.5}

        \fill[black!4, rounded corners=3pt] (\xoff, 0) rectangle (\xoff+13, -10);

        \begin{scope}[shift={(\xoff, 0)}]
          \draw[white, line width=1pt] (0, 0) grid[step=1] (13, -10);
        \end{scope}

        \draw[black!25, line width=0.8pt, rounded corners=3pt] (\xoff, 0) rectangle (\xoff+13, -10);

        \node[\col, font=\Large\bfseries, anchor=south] at (\xoff + 6.5, 0.2) {$\sym$};

        \foreach \k/\rl in {0/A, 1/2, 2/3, 3/4, 4/5, 5/6, 6/7, 7/8, 8/9, 9/{10}, 10/J, 11/Q, 12/K}{
          \node[font=\sffamily\tiny, text=black!60, anchor=north] at (\xoff + \k + 0.5, -10.1) {\rl};
        }
      }


      \pc{3}{2}{0}{cHand}
      \pc{0}{3}{0}{cHand}  \pc{0}{4}{0}{cHand}
      \pc{1}{2}{0}{cHand}  \pc{1}{9}{0}{cHand}  \pc{1}{10}{0}{cHand}  \pc{1}{11}{0}{cHand}
      \pc{2}{2}{0}{cHand}  \pc{2}{4}{0}{cHand}  \pc{2}{5}{0}{cHand}  \pc{2}{7}{0}{cHand}  \pc{2}{8}{0}{cHand}  \pc{2}{9}{0}{cHand}

      \pc{2}{0}{1}{cOpen}

      \pc{3}{9}{2}{cOwnDisc}

      \pc{2}{0}{3}{cOppDisc}

      \pc{3}{9}{4}{cOppPick}

      \pc{0}{5}{5}{cWild}  \pc{1}{5}{5}{cWild}  \pc{2}{5}{5}{cWild}  \pc{3}{5}{5}{cWild}

      \pc{3}{2}{6}{cSeen}  \pc{3}{4}{6}{cSeen}  \pc{3}{9}{6}{cSeen}
      \pc{0}{3}{6}{cSeen}  \pc{0}{4}{6}{cSeen}  \pc{0}{5}{6}{cSeen}
      \pc{1}{2}{6}{cSeen}  \pc{1}{9}{6}{cSeen}  \pc{1}{10}{6}{cSeen}  \pc{1}{11}{6}{cSeen}
      \pc{2}{0}{6}{cSeen}  \pc{2}{2}{6}{cSeen}  \pc{2}{4}{6}{cSeen}  \pc{2}{5}{6}{cSeen}  \pc{2}{7}{6}{cSeen}  \pc{2}{8}{6}{cSeen}  \pc{2}{9}{6}{cSeen}

      \pc{3}{2}{7}{cPureSet}  \pc{1}{2}{7}{cPureSet}  \pc{2}{2}{7}{cPureSet}
      \pc{1}{9}{7}{cImpureSeq}  \pc{1}{10}{7}{cImpureSeq}  \pc{1}{11}{7}{cImpureSeq}  \pc{2}{5}{7}{cImpureSeq}
      \pc{2}{7}{7}{cPureSeq}  \pc{2}{8}{7}{cPureSeq}  \pc{2}{9}{7}{cPureSeq}

      \pc{0}{2}{8}{cLive}  \pc{1}{4}{8}{cLive}

      \pc{0}{3}{9}{cPartial}  \pc{0}{4}{9}{cPartial}  \pc{2}{4}{9}{cPartial}

      \foreach \r/\lbl in {
        0/{(1) hand},
        1/{(2) open-top},
        2/{(3) own-discards},
        3/{(4) opp-discards},
        4/{(5) opp-picks},
        5/{(6) wildcard mask},
        6/{(7) seen-cards},
        7/{(8) meld-covered},
        8/{(9) live-outs},
        9/{(10) partial-melds}%
      }{
        \node[anchor=east, align=right, font=\sffamily\scriptsize, text=black!85]
        at (-0.5, {-\r - 0.5}) {\lbl};
      }

      \begin{scope}[shift={(0, -11.6)}]

        \draw[draw=black!30, fill=black!2, dashed, thick, rounded corners=4pt]
        (0, 0) rectangle (53.5, -2.8);

        \newcommand{\horizLegItem}[4]{
          \fill[#2, rounded corners=1.5pt] (#1, -0.9) rectangle (#1 + 1.0, -1.9);

          \node[anchor=north west, align=left, inner sep=0, font=\sffamily]
          at (#1 + 1.4, -0.9) {
            \textcolor{#2!85!black}{\textbf{#3}} \\[-0.1ex]
            \textcolor{black!75}{\footnotesize #4}
          };
        }

        \horizLegItem{1.5}{cPureSeq}{pure seq}{8\cs\,9\cs\,T\cs}
        \horizLegItem{15.0}{cImpureSeq}{impure seq}{T\ds\,J\ds\,Q\ds\,6\cs}
        \horizLegItem{28.5}{cPureSet}{pure set}{3\sps\,3\ds\,3\cs}
        \horizLegItem{42.0}{cPartial}{partial meld}{4\hs\,5\hs\,/\,5\hs\,5\cs}

        \node[anchor=north east, font=\sffamily\scriptsize\itshape, text=black!50]
        at (53.0, -0.6) {wildcard rank = 6};

      \end{scope}

    \end{tikzpicture}%
  }
  \caption{%
    Observation channel matrix for an example game state (wildcard rank\,$=$\,6).
    Each row is one of the ten binary channels; each column is one card in
    canonical suit$\,\times\,$rank order
    ($\varheartsuit$\,$\vardiamondsuit$\,$\clubsuit$\,$\spadesuit$, ranks A--K).
    A coloured cell is 1; light grey is 0.
    The agent's 13-card hand contains a
    \textcolor{cPureSeq!80!black}{\textbf{pure sequence}} (8\,9\,T\,$\clubsuit$),
    an \textcolor{cImpureSeq!90!black}{\textbf{impure sequence}}
    (T\,J\,Q\,$\vardiamondsuit$\,$+$\,6\,$\clubsuit$ wildcard),
    a \textcolor{cPureSet!80!black}{\textbf{pure set}} (3\,$\spadesuit\vardiamondsuit\clubsuit$),
    and two \textbf{partial melds} (4\,5\,$\varheartsuit$ and 5\,$\varheartsuit$5\,$\clubsuit$).
    Row~(8) cell colours identify the meld type; 6\,$\clubsuit$ appears in both
    row~(1) and row~(6) since it is a wildcard held in hand.
    Channels~(8)--(10) are computed online by the Numba-JIT meld analyser.%
  }
  \label{fig:obs}
\end{figure*}
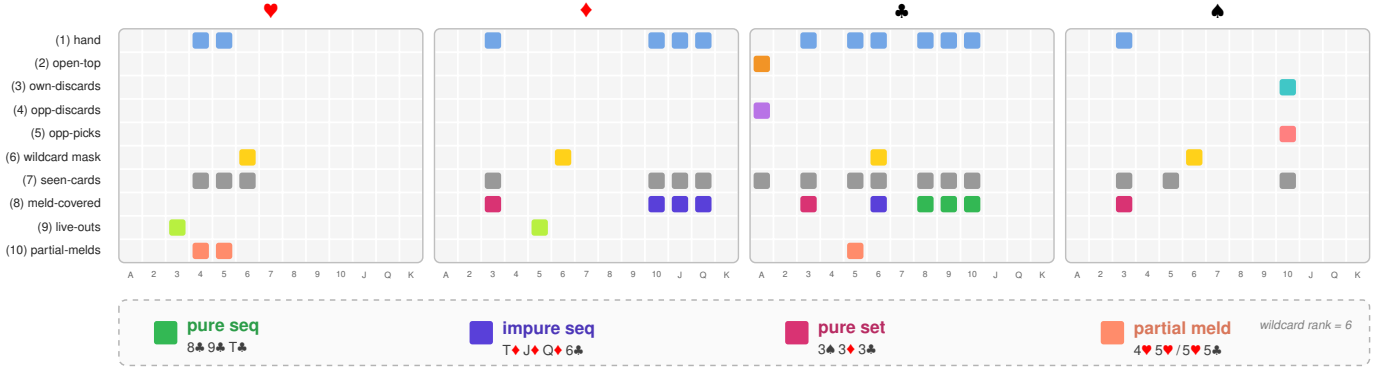

The agent's observation space is a 527-dimensional dense float32 vector, structured as a $10 \times 52$ binary feature matrix appended with seven global scalar features. The matrix encodes ten channels over the 52-card deck in canonical suit-rank order, as illustrated in Fig.~\ref{fig:obs}. We partition these channels into three categories:

\begin{itemize}
  \item Player and game history (Channels 1--5): the agent's current hand, the top card of the open discard pile, the agent's prior discards, the opponent's discards, and the opponent's historical picks from the open pile. This last channel serves as a mechanism for opponent-interest inference, as drawing from the open pile indicates a desire for a specific card.

  \item Game context (Channels 6--7): a static mask of all four cards matching the current game's wildcard rank, and a union of all seen cards defining the set of cards still unseen in the closed deck.

  \item Computed meld features (Channels 8--10): to directly overcome the meld-validity bottleneck described in §I, these channels are generated dynamically via a Numba-JIT (a just-in-time compiler for Python numerical code) meld analyser that equips the policy with explicit structural priors. They mark cards covered by a greedy maximum-coverage meld partition, highlight unseen cards that would immediately complete a partial meld, and flag cards belonging to a two-card partial meld. This computed block circumvents the need for explicit tree search by surfacing meld progress directly to the policy network at every step. The full analyser implementation is available in our open-source release.
\end{itemize}

Finally, the seven scalar features provide temporal and structural context. They include the fraction of the deck remaining, the normalised turn number, the current phase indicator, a boolean flag for the presence of a pure sequence, and three meld-summary statistics (number of complete melds, live outs, and partial melds) that give the policy a compact numerical snapshot of hand progress. The opponent's private hand and the exact composition of the closed deck remain hidden, meaning all opponent inferences must be derived from public discard and draw history.

\subsection{Rewards and Termination}

The environment yields a sparse terminal reward structure, providing $+1$ for a win, $-1$ for a loss, and $0$ for a draw. A win is achieved through a valid declaration, while a loss occurs if the opponent declares validly or if the agent executes an invalid declaration (though our action masking prevents the RL policy from encountering this terminal state during training). Draws are triggered by the exhaustion of the closed deck or by reaching a 100-turn limit. Because intermediate actions lack intrinsic value, we augment the environment during training with a potential-based shaping reward applied at every draw and non-terminal discard step. This shaping signal densifies the sparse reward without altering the set of optimal policies, as detailed in §\ref{sec:shaping}.

\section{Method}

\subsection{Policy Architecture}


\definecolor{cArchChan}  {rgb}{0.45,0.65,0.90}   
\definecolor{cArchScal}  {rgb}{1.00,0.82,0.10}   
\definecolor{cArchSeq}   {rgb}{0.20,0.72,0.33}   
\definecolor{cArchSet}   {rgb}{0.95,0.58,0.15}   
\definecolor{cArchTrunk} {rgb}{0.25,0.78,0.78}   
\definecolor{cArchPol}   {rgb}{0.72,0.45,0.90}   
\definecolor{cArchVal}   {rgb}{1.00,0.50,0.50}   

\begin{figure}[t]
  \centering
  \resizebox{\linewidth}{!}{%
    \begin{tikzpicture}[
        >=stealth,
        font=\sffamily\scriptsize,
        thick,
        box/.style={rectangle, rounded corners=3pt, align=center, inner sep=4pt, draw=black!60},
        chan/.style={box, fill=cArchChan!15, draw=cArchChan!90, text=black!90},
        scal/.style={box, fill=cArchScal!20, draw=cArchScal!90, text=black!90, minimum height=0.6cm},
        seq/.style={box, fill=cArchSeq!15, draw=cArchSeq!90, text=black!90},
        set/.style={box, fill=cArchSet!15, draw=cArchSet!90, text=black!90},
        trunk/.style={box, fill=cArchTrunk!15, draw=cArchTrunk!90, text=black!90, minimum height=1.5cm},
        pol/.style={box, fill=cArchPol!15, draw=cArchPol!90, text=black!90},
        val/.style={box, fill=cArchVal!15, draw=cArchVal!90, text=black!90},
        concat/.style={rectangle, rounded corners=2pt, draw=black!50, fill=black!10, minimum width=0.4cm, minimum height=3.6cm, align=center, text=black!70},
        arrow/.style={->, draw=black!50, thick, rounded corners=2pt}
      ]

      \coordinate (inCenter) at (0,0);
      \foreach \i in {2,1} {
        \node[chan, minimum width=1.5cm, minimum height=2.2cm, shift={(\i*0.1,\i*0.1)}] at (inCenter) {};
      }
      \node[chan, minimum width=1.5cm, minimum height=2.2cm] (input) at (inCenter) {\textbf{Input}\\$10{\times}4{\times}13$};

      \node[scal, below=0.3cm of input] (scalars) {$+\ 7$ scalars};

      \node[seq, right=0.7cm of input, yshift=1.0cm] (bSeq) {\textbf{Conv2d}\\$1{\times}3$, 32 filters\\ReLU + max-suit\\\textit{seq branch}};
      \node[set, right=0.7cm of input, yshift=-1.0cm] (bSet) {\textbf{Conv2d}\\$4{\times}1$, 32 filters\\ReLU\\\textit{set branch}};

      \node[concat, right=0.6cm of bSeq, yshift=-1.0cm] (concat) {\rotatebox{90}{\textbf{concat}}};

      \node[trunk, right=0.6cm of concat] (fc512) {\textbf{FC 512}\\LayerNorm\\ReLU};
      \node[trunk, right=0.4cm of fc512] (fc256) {\textbf{FC 256}\\LayerNorm\\ReLU};

      \node[pol, right=0.8cm of fc256, yshift=1.0cm] (hPol) {\textbf{Policy head}\\FC $\to$ 55\\mask $-\infty$, softmax};
      \node[val, right=0.8cm of fc256, yshift=-1.0cm] (hVal) {\textbf{Value head}\\FC $\to$ 1};

      \node[box, right=0.4cm of hPol, font=\sffamily\small, fill=white] (outPol) {$\pi(a{\mid}o)$};
      \node[box, right=0.4cm of hVal, font=\sffamily\small, fill=white] (outVal) {$V(o)$};

      \draw[arrow] (input.east) -- ++(0.4,0) |- (bSeq.west);
      \draw[arrow] (input.east) -- ++(0.4,0) |- (bSet.west);

      \draw[arrow] (bSeq.east) -- (bSeq.east -| concat.west) node[midway, above, font=\tiny, text=black!60] {416};
      \draw[arrow] (bSet.east) -- (bSet.east -| concat.west) node[midway, above, font=\tiny, text=black!60] {416};

      \path (concat.west) ++(-0.3cm, 0) coordinate (preConcat);
      \draw[arrow] (scalars.east) -- (scalars.east -| preConcat) |- ([yshift=-1.4cm]concat.west);

      \draw[arrow] (concat.east) -- (fc512.west) node[midway, above, font=\tiny, text=black!60] {839};
      \draw[arrow] (fc512.east) -- (fc256.west);

      \draw[arrow] (fc256.east) -- ++(0.5,0) |- (hPol.west);
      \draw[arrow] (fc256.east) -- ++(0.5,0) |- (hVal.west);

      \draw[arrow] (hPol.east) -- (outPol.west);
      \draw[arrow] (hVal.east) -- (outVal.west);

      \begin{scope}[on background layer]
        \node[rectangle, draw=black!30, fill=cArchTrunk!5, dashed, rounded corners=4pt, inner sep=8pt,
        fit=(fc512)(fc256), label={[font=\sffamily\tiny\itshape, text=black!50, yshift=-1mm]below:shared trunk}] (trunkbox) {};
      \end{scope}

      \coordinate (legStart) at ([xshift=-1cm, yshift=-1.5cm]trunkbox.south west);
      \begin{scope}[every node/.style={font=\sffamily\tiny, right, text=black!80}]
        \node[chan, inner sep=0, minimum width=4mm, minimum height=2.5mm] (leg1) at (legStart) {}; \node at ([xshift=0.5mm]leg1.east) (t1) {input};
        \node[seq,  inner sep=0, minimum width=4mm, minimum height=2.5mm] (leg2) at ([xshift=2mm]t1.east) {}; \node at ([xshift=0.5mm]leg2.east) (t2) {seq branch};
        \node[set,  inner sep=0, minimum width=4mm, minimum height=2.5mm] (leg3) at ([xshift=2mm]t2.east) {}; \node at ([xshift=0.5mm]leg3.east) (t3) {set branch};
        \node[trunk,inner sep=0, minimum width=4mm, minimum height=2.5mm] (leg4) at ([xshift=2mm]t3.east) {}; \node at ([xshift=0.5mm]leg4.east) (t4) {trunk};
        \node[pol,  inner sep=0, minimum width=4mm, minimum height=2.5mm] (leg5) at ([xshift=2mm]t4.east) {}; \node at ([xshift=0.5mm]leg5.east) (t5) {policy};
        \node[val,  inner sep=0, minimum width=4mm, minimum height=2.5mm] (leg6) at ([xshift=2mm]t5.east) {}; \node at ([xshift=0.5mm]leg6.east) (t6) {value};
      \end{scope}

    \end{tikzpicture}%
  }
  \caption{%
    IRumAI dual-branch network architecture ($\sim$580K parameters). The $10 \times 4 \times 13$ observation tensor is processed by two specialised convolutional paths: a $1{\times}3$ sequence branch that slides horizontally across ranks to detect consecutive patterns, and a $4{\times}1$ set branch that slides vertically across suits to aggregate matching ranks.%
  }
  \label{fig:arch}
\end{figure}
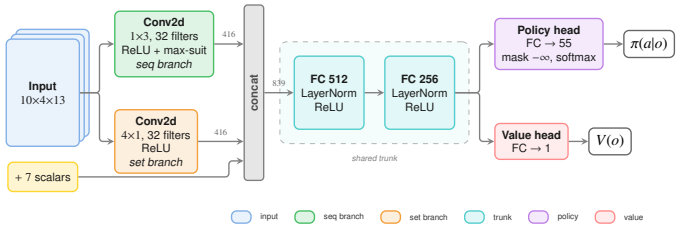

To process the ten-channel observation tensor, we designed a dual-branch convolutional architecture tailored to the combinatorial structure of Indian Rummy, as illustrated in Fig.~\ref{fig:arch}. The network's inductive bias aligns with the game's two valid meld types: sequences, which require rank-adjacency within a single suit, and sets, which require same-rank agreement across multiple suits. The first branch, the sequence branch, employs a $1{\times}3$ convolution with 32 filters and same-padding along the rank axis, sliding horizontally across the rank dimension to detect adjacent sequence patterns. This is followed by a ReLU activation and a max-pooling operation over the suit dimension, allowing the network to extract the strongest sequence pattern at each rank position regardless of the suit in which it appears. The second branch, the set branch, uses a $4{\times}1$ convolution with 32 filters and valid padding, sliding vertically across the suit dimension to aggregate evidence of matching ranks and detect potential sets. This is followed by a ReLU activation.

Both branches produce 416-dimensional feature vectors: the sequence branch yields $32\text{ filters}\times13\text{ ranks}=416$ after max-pooling collapses the suit dimension, while the set branch yields $32\text{ filters}\times13\text{ ranks}=416$ after valid-padding collapses the suit dimension to size 1. These are concatenated with the seven global scalar features, yielding an $416+416+7=839$-dimensional fused representation. This representation is processed by a shared two-layer fully connected trunk (512 and 256 units, each followed by LayerNorm and ReLU). The trunk learns higher-order interactions between the board state and the current game phase, after which its output splits into a scalar value estimator and a 55-dimensional policy logit vector. To ensure phase-correct action selection, we apply legal-action masking by setting the logits of all illegal actions to a large negative value prior to the softmax activation, zeroing out invalid probabilities while remaining numerically stable for half-precision training.

\subsection{PPO Optimisation}

We optimise the policy using Proximal Policy Optimization (PPO)~\cite{schulman2017ppo} with Generalized Advantage Estimation (GAE)~\cite{schulman2016gae}. Table~\ref{tab:hparams} summarises the full hyperparameter configuration. All hidden linear and convolutional layers use orthogonal initialisation with a gain of $\sqrt{2}$, while the policy head is initialised with a gain of $0.01$ to ensure a near-uniform initial action distribution. Our PPO hyperparameters mostly follow the defaults of Schulman et al. The main exception is the entropy coefficient, which we raised from the more typical $0.01$ to $0.08$ after preliminary runs at lower values collapsed onto a small subset of the 13 near-equivalent discard candidates available per turn. The value-loss coefficient, target-KL early-stopping threshold of $0.015$, and batch sizing were tuned in the same pass for stability and to suit the GH200's parallel-env throughput.

\begin{table}[t]
  \centering
  \caption{PPO hyperparameters.}
  \label{tab:hparams}
  \begin{tabular}{lr}
    \toprule
    Hyperparameter & Value \\
    \midrule
    Discount factor $\gamma$ & 0.99 \\
    GAE $\lambda$ & 0.95 \\
    Clip range $\epsilon_{\text{clip}}$ & 0.2 \\
    Entropy coefficient & 0.08 \\
    Value-function coefficient & 0.25 \\
    Target KL & 0.015 \\
    Max gradient norm & 0.5 \\
    Learning rate & $3{\times}10^{-4} \to 3{\times}10^{-5}$ (cosine) \\
    LR warmup & 50 updates \\
    Parallel environments & 128 \\
    Rollout length & 256 steps/env \\
    Batch size & 32\,768 \\
    Mini-batch size & 8\,192 \\
    \bottomrule
  \end{tabular}
\end{table}

\subsection{Reward Shaping}
\label{sec:shaping}

The native environmental reward for Indian Rummy is sparse, consisting of a $\{-1, 0, +1\}$ terminal signal. To provide the agent with a dense learning gradient, we implement potential-based reward shaping\footnote{Although the deadwood potential derives from the game's native scoring, making it arguably more natural than typical hand-crafted shaping functions, we retain the term because the signal formally satisfies the Ng~et~al.\ potential-difference framework and is applied as an augmentation to the sparse terminal reward, not as a replacement.} at every draw and non-terminal discard step. Following the theoretical framework of Ng~et~al.~\cite{ng1999shaping}, we transform the native reward $R(s, a, s')$ into a dense reward $R'(s, a, s') = R(s, a, s') + F(s, s')$, where the shaping term $F$ is defined as a difference in state potentials:
\begin{equation}
  F(s, s') = \Phi(s') - \Phi(s)
\end{equation}
The potential function $\Phi(s)$ is grounded in the game's deadwood scoring rule:
\begin{equation}
  \Phi(s) = -\frac{\text{meld\_deadwood}(h)}{130}
\end{equation}
The term $\text{meld\_deadwood}(h)$ represents the sum of the point values of all cards not covered by a complete meld in a greedy maximum-coverage partition of the agent's hand $h$. Because the maximum possible deadwood in a 13-card hand is 130 points, this function bounds the state potential to $[-1, 0]$.

By tying the potential to the rules of the game, we provide the agent with step-by-step feedback that reinforces actions incrementally completing melds. Because $F(s, s')$ takes the form of a potential difference, the accumulated shaping reward along any sequence of actions from state $s_0$ to a terminal state $s_T$ strictly telescopes to $\Phi(s_T) - \Phi(s_0)$. This mathematical property guarantees that the set of optimal policies for the shaped Markov Decision Process remains identical to that of the unshaped environment, allowing us to safely densify the learning gradient without altering optimal-policy behaviour.

\subsection{Training Opponents}
\label{sec:opponents}

Training is initialised from a behaviour-cloning (BC) checkpoint. We generate 10{,}000 games of MinDistOppAgent self-play, extract state--action pairs from both players, and train the policy network to minimise cross-entropy against the expert's actions for up to 20 epochs (Adam, lr\,$=$\,$3{\times}10^{-4}$, batch size 4{,}096), with early stopping once action-match accuracy exceeds 80\%. This warm start bypasses the initial random-exploration phase and grounds the policy in a competent baseline of rummy play. To prevent the RL policy from drifting too far from this initialisation during early training, we add a KL-divergence penalty toward the frozen BC policy with coefficient $0.05$, linearly decayed to zero over the first 500 PPO updates.

For our primary model, the training opponent pool is kept fixed and consists of the MinScoreAgent and MinScoreOppAgent baselines. We deliberately restrict the training pool to these faster, simpler agents to demonstrate that the agent can learn a highly generalisable policy from cheap heuristic opponents, transferring to the more complex, search-based MinDist family during evaluation without any explicit RL-time exposure. We note, however, that the BC warm-start uses MinDistOppAgent self-play data, so the policy has indirect exposure to MinDist-style play patterns prior to RL optimisation.

\section{Experimental Setup}

\subsection{Training Protocol}

All experiments were conducted on a single NVIDIA GH200 GPU using CUDA acceleration and automatic mixed precision (AMP with float16 forward passes and float32 gradient accumulation). Training was executed for 10{,}000 updates, corresponding to roughly 327.7 million environment steps, which took approximately 41 hours of wall-clock time at a processing rate of 2{,}400 steps per second. To ensure resilience, checkpoints were saved every 200 updates, while the agent's progress was monitored every 100 updates via in-training tournaments of 100 games per opponent. For the final held-out evaluation, we ran 1{,}000 games per opponent across five random seeds, yielding 5{,}000 games per matchup.

\subsection{Metrics}

The primary metric is tournament win rate against each opponent. To contextualise the RL agent's skill level, we additionally report the inter-heuristic win-rate matrix we obtain by running our reimplementations of the heuristics from~\cite{saha2025rummy}, enabling direct comparison of IRumAI against the heuristic skill ladder.

\section{Results}
\label{sec:results}

\begin{table*}[t]
  \centering
  \caption{Tournament win probabilities against heuristic baselines. The left-hand matrix (Baselines vs.\ Opponent) establishes a reference skill ladder, evaluated using our implementation of the agents introduced in~\cite{saha2025rummy}. All values are shown as mean win probabilities. \textbf{Bold} indicates the highest win rate in each row.}  \label{tab:unified_results}
  \renewcommand{\arraystretch}{1.2}
  \setlength{\tabcolsep}{4pt}
  \begin{tabular}{@{} l rrrrrr c rcrrr @{}}
    \toprule
    \multirow{2}{*}{\textbf{Opponent}} & \multicolumn{6}{c}{\textbf{Baseline Win Rate vs.\ Opponent~\cite{saha2025rummy}}} && \multicolumn{5}{c}{\textbf{IRumAI (Ours) vs.\ Opponent}} \\
    \cmidrule{2-7} \cmidrule{9-13}
    & Rnd & MinSc & MinScO & MinDi & MinDiS & MinDiO && Win & CI\textsubscript{95} & FM & SM & $\hat{a}$ \\
    \midrule
    RandomAgent       & - & 0.955 & 0.941 & 0.997 & 0.995 & 0.997 && \textbf{0.9983$\pm$0.0006} & $\pm$0.0025 & 0.9980$\pm$0.0020 & 0.9987$\pm$0.0012 & -0.0003 \\
    MinScoreAgent     & 0.005 & - & 0.509 & 0.581 & 0.610 & 0.604 && \textbf{0.6692$\pm$0.0112} & $\pm$0.0291 & 0.7140$\pm$0.0131 & 0.6227$\pm$0.0110 & 0.0457 \\
    MinScoreOppAgent  & 0.003 & 0.442 & - & 0.599 & 0.604 & 0.603 && \textbf{0.6847$\pm$0.0101} & $\pm$0.0287 & 0.7493$\pm$0.0110 & 0.6173$\pm$0.0115 & 0.0660 \\
    MinDistAgent      & 0.003 & 0.410 & 0.384 & - & 0.509 & 0.510 && \textbf{0.5715$\pm$0.0088} & $\pm$0.0306 & 0.6007$\pm$0.0070 & 0.5400$\pm$0.0120 & 0.0303 \\
    MinDistScoreAgent & 0.001 & 0.381 & 0.383 & 0.489 & - & 0.501 && \textbf{0.5565$\pm$0.0031} & $\pm$0.0308 & 0.5993$\pm$0.0042 & 0.5127$\pm$0.0050 & 0.0433 \\
    MinDistOppAgent   & 0.003 & 0.384 & 0.381 & 0.482 & 0.495 & - && \textbf{0.5387$\pm$0.0118} & $\pm$0.0309 & 0.5707$\pm$0.0151 & 0.5040$\pm$0.0092 & 0.0333 \\
    \bottomrule
  \end{tabular}
\end{table*}

\begin{table}[t]
  \centering
  \caption{Mean per-action inference latency. Values are shown in milliseconds.}
  \label{tab:latency}
  \begin{tabular}{lr}
    \toprule
    Agent & Latency (ms) \\
    \midrule
    RandomAgent       & $0.05$ \\
    MinScoreAgent     & $26$ \\
    MinScoreOppAgent  & $40$ \\
    MinDistAgent      & $2403$ \\
    MinDistScoreAgent & $2394$ \\
    MinDistOppAgent   & $2426$ \\
    \midrule
    IRumAI (ours) & $0.33$ \\
    \bottomrule
  \end{tabular}
\end{table}

Table~\ref{tab:unified_results} reports the held-out tournament results at update 10{,}000. All figures are mean win rates over the five evaluation seeds. Against RandomAgent, IRumAI achieves a near-perfect mean win rate of 99.83\%. Performance stays strong against the MinScore family, winning 66.92\% of games against MinScoreAgent and 68.47\% against MinScoreOppAgent. These margins indicate that the agent has learned the core combinatorial requirement of partitioning thirteen cards into valid melds.

A harder test of the policy's strategic depth is its performance against the \textit{MinDist} agents, which rely on exhaustive combinatorial search to minimise edit distance. None of these agents were included in the training pool, so the agent's performance here measures transfer to opponents unseen during RL training. IRumAI achieves mean win rates of 57.15\%, 55.65\%, and 53.87\% against MinDistAgent, MinDistScoreAgent, and MinDistOppAgent, respectively. Even against MinDistOppAgent, which incorporates explicit opponent modelling via search, our agent maintains a winning edge while using significantly less computation.

Comparing these results to the baseline heuristics puts IRumAI's performance into perspective. In the inter-heuristic evaluation, MinDistOppAgent achieves about a 60\% win rate against both MinScore agents. IRumAI wins approximately 67\% and 68\% of games against MinScoreAgent and MinScoreOppAgent respectively, exceeding the strongest heuristic's margin. Against the MinDist family, where inter-heuristic matchups cluster near 50\%, IRumAI keeps a consistent lead. This suggests the neural policy learned a robust strategy for navigating the hidden-information state space without relying on explicit lookahead.

Positional bias is a known factor in Indian Rummy, where the player moving first often has an advantage. Following Saha~et~al.~\cite{saha2025rummy}, we decompose the overall win probability into first-mover ($p_\text{FM}$) and second-mover ($p_\text{SM}$) components and estimate the first-mover advantage as $\hat{a} = (p_\text{FM} - p_\text{SM})/2$. As Table~\ref{tab:unified_results} shows, IRumAI exhibits an advantage ranging from 3.0 to 6.6 percentage points across opponents. This aligns with the 4 to 6 percentage point range seen in inter-heuristic matchups and reported in prior work. Because the evaluation alternates the starting player, the overall win rates reflect actual skill independent of positional luck.

A major practical benefit of the learned policy is its speed. As Table~\ref{tab:latency} details, the search-based MinDist heuristics take about 2.4 seconds per action due to their exhaustive enumeration. In contrast, IRumAI requires only a single forward pass through its 580{,}000-parameter network, taking 0.33 milliseconds per action at batch size 1 on a single ARM Neoverse-V2 core with no GPU. This figure measures the policy forward pass and masked argmax decode. Observation construction is excluded because it is shared with the heuristic baselines for fairness. This is at least $79\times$ faster than the MinScore family and over $7{,}000\times$ faster than the MinDist baselines, placing the agent well within real-time latency requirements.

\section{Analysis}

\subsection{Ablation Studies}
\label{sec:ablations}

Having established that IRumAI transfers to unseen opponents and achieves significant inference speedups against heuristic baselines, we now dissect the specific training components that enabled this performance. We conducted four ablation studies to understand the impact of reward shaping, behaviour-cloning (BC) initialisation, curriculum scheduling, and policy architecture. Each ablation reports a single seed, as compute constraints precluded full multi-seed runs.

First, we tested the potential-based reward shaping by training an agent using only the sparse terminal rewards. Table~\ref{tab:shaping_ablation} shows that dropping the meld-aware shaping leads to a performance drop of up to 4.9 percentage points across the opponent pool. Because this shaping uses the game's native deadwood scoring and mathematically preserves the optimal policy, it provides a consistent performance boost without altering optimal-policy behaviour.

\begin{figure*}[t]
  \centering
  \includegraphics[width=\linewidth]{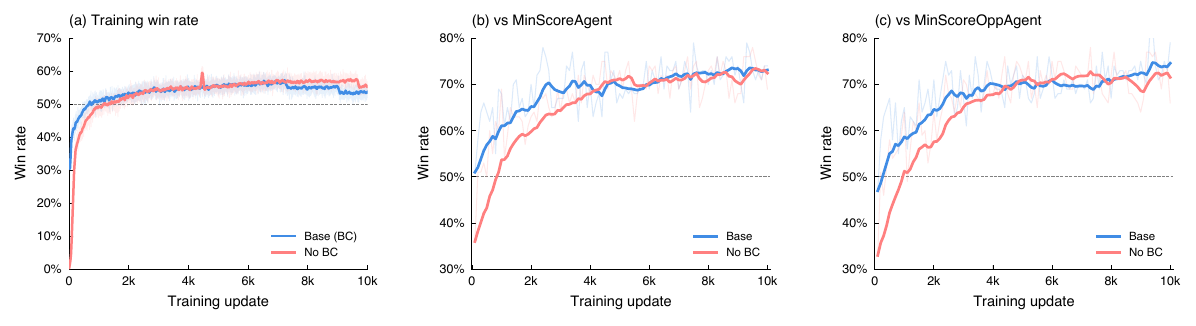}
  \caption{Behaviour-cloning ablation. (a) Smoothed training win rate against the opponent pool over 10{,}000 updates (raw trace at reduced opacity). (b, c) Periodic held-out evaluation win rates against MinScoreAgent and MinScoreOppAgent. The BC-initialised run (blue) maintains a clear early lead before converging with the cold-start run beyond $\sim$7{,}000 updates.}
  \label{fig:ablation_curves}
\end{figure*}

Next, we evaluated the BC warm-start by training an agent from scratch while keeping everything else identical. As Fig.~\ref{fig:ablation_curves} shows across its three panels, the BC-initialised run leads the cold-start agent by up to 15 percentage points in mean win rate during the first 2{,}000 updates. However, by update 7{,}000, both runs converge to comparable performance. This confirms that BC accelerates early learning but is not strictly necessary for final policy quality. This early lag in the cold-start curve likely reflects the sparse terminal reward signal under strict declaration masking. Random play almost never assembles a valid 13-card partition, so the agent must accumulate enough potential-based shaping signal before win rate begins to climb above the heuristic floor.

We also experimented with a Prioritised Fictitious Self-Play (PFSP) curriculum~\cite{vinyals2019alphastar}. As detailed in Table~\ref{tab:pfsp_ablation}, PFSP yielded inconsistent results: it gained 3.4 percentage points against MinScoreOppAgent but lost 3.7 points against MinDistOppAgent, with no consistent improvement over uniform opponent sampling. Post-training PFSP fine-tuning showed similarly mixed results.

We attribute this to two structural properties of the training setup. First, the opponent pool lacks non-transitive dynamics: it forms a strict skill ladder (Random $<$ MinScore $<$ MinDist) without the rock-paper-scissors cycles PFSP was designed to handle~\cite{vinyals2019alphastar}, so dynamically re-weighting a monotonic hierarchy offers no measurable advantage over uniform sampling. Second, the two training opponents share the same core greedy-deadwood-minimisation logic, differing only in a discard-safety tiebreaker. Re-weighting between near-identical strategies provides little curriculum signal, so the priority weights end up tracking estimation noise rather than meaningful strategic diversity.

Finally, to isolate the impact of our architectural design, we replaced the dual-branch convolutional policy with a vanilla MLP of identical trunk width, trained under the same PPO configuration for the full 10{,}000 updates. As Table~\ref{tab:shaping_ablation} shows, the MLP trails the convolutional agent by 8 to 11 percentage points across all heuristic opponents. Most notably, it falls below a winning threshold against MinDistOppAgent (43.2\% vs.\ 53.9\%). This confirms that the sequence- and set-based inductive biases are essential drivers of the agent's performance, rather than generic PPO optimisation.

\begin{table}[t]
  \centering
  \caption{Reward-shaping, behaviour-cloning, and architecture ablations.}
  \label{tab:shaping_ablation}
  \begin{tabular}{lrrrr}
    \toprule
    Opponent & Base & \makecell{No\\Shaping} & \makecell{No\\BC} & \makecell{Vanilla\\MLP} \\
    \midrule
    RandomAgent       & \textbf{0.9983} & 0.9990 & 0.9900 & 0.9930 \\
    MinScoreAgent     & \textbf{0.6692} & 0.6390 & 0.6461 & 0.5780 \\
    MinScoreOppAgent  & \textbf{0.6847} & 0.6360 & 0.6770 & 0.5970 \\
    MinDistAgent      & \textbf{0.5715} & 0.5470 & 0.5304 & 0.4820 \\
    MinDistScoreAgent & \textbf{0.5565} & 0.5290 & 0.5470 & 0.4600 \\
    MinDistOppAgent   & \textbf{0.5387} & 0.5220 & 0.5242 & 0.4320 \\
    \bottomrule
  \end{tabular}
\end{table}

\begin{table}[t]
  \centering
  \caption{Prioritised Fictitious Self-Play (PFSP) curriculum ablation.}
  \label{tab:pfsp_ablation}
  \begin{tabular}{lrrr}
    \toprule
    Opponent & Uniform & PFSP & \makecell{Post-tr.\\PFSP} \\
    \midrule
    RandomAgent       & \textbf{0.9983} & 0.9980 & 0.9990 \\
    MinScoreAgent     & 0.6692 & \textbf{0.6760} & 0.6480 \\
    MinScoreOppAgent  & 0.6847 & \textbf{0.7190} & 0.6940 \\
    MinDistAgent      & \textbf{0.5715} & 0.5740 & 0.5470 \\
    MinDistScoreAgent & 0.5565 & \textbf{0.5630} & 0.5320 \\
    MinDistOppAgent   & \textbf{0.5387} & 0.5020 & 0.5270 \\
    \bottomrule
  \end{tabular}
\end{table}

\subsection{Emergent Opponent Modelling}
\label{sec:probe}

Finally, to determine whether the policy network implicitly tracks the opponent's hidden cards, we trained a linear probe on the policy's frozen 256-dimensional trunk activations. We recorded the trunk representation and the ground-truth opponent hand at every RL-agent decision step (both draw and discard nodes) across 6{,}000 evaluation games (2{,}000 games each against MinScoreAgent, MinScoreOppAgent, and MinDistOppAgent), using a game-level 80/20 random train/validation split to prevent information leakage between observations from the same game. We then fit a single linear layer to predict the 52-dimensional binary hand vector.

We compared this against a card-counting baseline that distributes probability uniformly over all cards not known to be elsewhere (i.e.\ not in the agent's hand, the discard pile, or the open pile). As shown in Table~\ref{tab:probe}, the linear probe outperforms card counting, achieving an AUC-ROC of 0.744 and a top-13 recall of 47.2\%. While the AUC improvement is modest (3.4 percentage points), the 9.9 percentage point increase in top-13 recall is more revealing as it shows that the trunk activations are substantially better at identifying the specific cards most likely to be in the opponent's hand, beyond what public information alone can determine. This suggests the network has learned to extract predictive signal from the opponent's interaction history, such as which cards they drew from the open pile and which they chose to discard.

\begin{table}[t]
  \centering
  \caption{Opponent-hand prediction performance from frozen trunk activations.}
  \label{tab:probe}
  \begin{tabular}{lccc}
    \toprule
    Method & AUC-ROC & Avg Prec & Top-13 Recall \\
    \midrule
    Random            & 0.500 & 0.250 & 0.249 \\
    Unseen-uniform    & 0.710 & 0.410 & 0.373 \\
    Linear probe      & \textbf{0.744} & \textbf{0.481} & \textbf{0.472} \\
    \bottomrule
  \end{tabular}
\end{table}

\section{Discussion and Conclusion}

We introduced IRumAI, the first reinforcement learning agent for Indian Rummy. By pairing a PPO policy with a meld-aware observation encoding and deadwood-based reward shaping, the agent learns to navigate the game's large combinatorial state space without any explicit search. It maintains a consistent winning edge against the full heuristic hierarchy while requiring only 0.33\,ms per action, enabling real-time deployment on commodity hardware.

Our ablation studies reveal a clear hierarchy among design choices. The dual-branch convolutional architecture is the single largest contributor, providing an 8 to 11 percentage point edge over a vanilla MLP and proving decisive for achieving a positive win rate against the strongest search-based opponents. Potential-based reward shaping delivers a consistent boost of up to 5 percentage points across the opponent pool without altering optimal-policy behaviour, while behaviour-cloning initialisation accelerates convergence by several thousand updates without materially affecting final performance. Prioritised Fictitious Self-Play failed to improve over uniform opponent sampling, which we attribute to the lack of non-transitivity in our opponent pool (§\ref{sec:ablations}). Beyond win rates, linear probing reveals that the policy trunk encodes implicit estimates of the opponent's hidden hand, suggesting that the agent has learned to infer private information from the observable interaction history rather than relying solely on its own hand optimisation.

Several limitations bound these conclusions. All ablation runs use a single seed, so reported differences of a few percentage points may partly reflect seed variance. The training opponent pool is intentionally narrow (only two heuristic agents), and while this supports our transfer narrative, it leaves open whether the policy would generalise to qualitatively different adversaries, particularly human players, whom we have not evaluated against. Finally, all experiments target the two-player, single-round format. The agent's behaviour in multi-player or multi-round variants, where long-term point management across rounds arises, remains untested.

These limitations suggest natural directions for future work. Expanding the training pool to include human demonstration data, and running head-to-head matches against amateur and expert human players, would test whether the policy can absorb richer strategic diversity and hold up against the opponents it will ultimately face. Integrating lightweight inference-time search, such as Monte Carlo rollouts conditioned on the trunk's opponent-hand estimates, could raise the tactical ceiling by catching subtle declaration opportunities that a pure policy may miss. This would combine the speed of the learned policy with the tactical depth of search rather than treating the two as alternatives. Scaling to multi-player formats like Pool Rummy will require extending the observation encoding to track multiple independent discard histories and long-horizon point accumulation. We publicly release our environment and training codebase to support further research in this under-explored domain at \href{https://github.com/vdesmond/IRumAI}{https://github.com/vdesmond/IRumAI}.

\section*{Acknowledgements}

The author thanks EURECOM for financial support enabling presentation of this work at IEEE CoG 2026, as well as for providing access to the NVIDIA GH200 GPU used for all experiments.

\bibliographystyle{IEEEtran}
\bibliography{refs}

\begin{thebibliography}{10}
\providecommand{\url}[1]{#1}
\csname url@samestyle\endcsname
\providecommand{\newblock}{\relax}
\providecommand{\bibinfo}[2]{#2}
\providecommand{\BIBentrySTDinterwordspacing}{\spaceskip=0pt\relax}
\providecommand{\BIBentryALTinterwordstretchfactor}{4}
\providecommand{\BIBentryALTinterwordspacing}{\spaceskip=\fontdimen2\font plus
\BIBentryALTinterwordstretchfactor\fontdimen3\font minus
  \fontdimen4\font\relax}
\providecommand{\BIBforeignlanguage}[2]{{%
\expandafter\ifx\csname l@#1\endcsname\relax
\typeout{** WARNING: IEEEtran.bst: No hyphenation pattern has been}%
\typeout{** loaded for the language `#1'. Using the pattern for}%
\typeout{** the default language instead.}%
\else
\language=\csname l@#1\endcsname
\fi
#2}}
\providecommand{\BIBdecl}{\relax}
\BIBdecl

\bibitem{moravcik2017deepstack}
M.~Morav{\v{c}}{\'\i}k, M.~Schmid, N.~Burch, V.~Lis{\'y}, D.~Morrill, N.~Bard,
  T.~Miller, K.~Waugh, M.~Johanson, and M.~Bowling, ``{DeepStack}: Expert-level
  artificial intelligence in heads-up no-limit poker,'' \emph{Science}, vol.
  356, no. 6337, pp. 508--513, 2017.

\bibitem{saha2025rummy}
\BIBentryALTinterwordspacing
P.~Saha, A.~Chakraborty, S.~Sarkar, S.~Maitra, D.~Mukherjee, and T.~Mukherjee,
  ``Quantitative rule-based strategy modeling in classic {Indian Rummy}: {A}
  metric optimization approach,'' 2025. [Online]. Available:
  \url{https://arxiv.org/abs/2601.00024}
\BIBentrySTDinterwordspacing

\bibitem{goldman2021mmd}
P.~Goldman, C.~R. Knutson, R.~Mahtab, J.~Maloney, J.~B. Mueller, and R.~G.
  Freedman, ``Evaluating gin rummy hands using opponent modeling and myopic
  meld distance,'' in \emph{Proceedings of the Thirty-Fifth {AAAI} Conference
  on Artificial Intelligence ({AAAI-21})}.\hskip 1em plus 0.5em minus
  0.4em\relax {AAAI} Press, 2021, pp. 14\,965--14\,966.

\bibitem{eicholtz2021heisenbot}
M.~Eicholtz, S.~Moss, M.~Traino, and C.~Roberson, ``Heisenbot: A rule-based
  game agent for gin rummy,'' in \emph{Proceedings of the Thirty-Fifth {AAAI}
  Conference on Artificial Intelligence ({AAAI-21})}.\hskip 1em plus 0.5em
  minus 0.4em\relax {AAAI} Press, 2021, pp. 15\,489--15\,495.

\bibitem{gallucci2021fitness}
J.~Gallucci, R.~Bowser, S.~Kettell, and C.~Overton, ``Estimating card fitness
  for discard in gin rummy,'' in \emph{Proceedings of the Thirty-Fifth {AAAI}
  Conference on Artificial Intelligence ({AAAI-21})}.\hskip 1em plus 0.5em
  minus 0.4em\relax {AAAI} Press, 2021, pp. 15\,503--15\,509.

\bibitem{nguyen2021ginnn}
V.~D. Nguyen, D.~Doan, and T.~W. Neller, ``A deterministic neural network
  approach to playing gin rummy,'' in \emph{Proceedings of the Thirty-Fifth
  {AAAI} Conference on Artificial Intelligence ({AAAI-21})}.\hskip 1em plus
  0.5em minus 0.4em\relax {AAAI} Press, 2021, pp. 14\,967--14\,968.

\bibitem{eswaran2020gaim}
S.~Eswaran, V.~Vimal, D.~Seth, and T.~Mukherjee, ``{GAIM}: Game action
  information mining framework for multiplayer online card games (rummy as case
  study),'' in \emph{Advances in Knowledge Discovery and Data Mining
  ({PAKDD})}, ser. Lecture Notes in Computer Science, vol. 12085.\hskip 1em
  plus 0.5em minus 0.4em\relax Springer, 2020, pp. 435--448.

\bibitem{zha2021douzero}
D.~Zha, J.~Xie, W.~Ma, S.~Zhang, X.~Lian, X.~Hu, and J.~Liu, ``{DouZero}:
  Mastering {DouDizhu} with self-play deep reinforcement learning,'' in
  \emph{Proceedings of the 38th International Conference on Machine Learning
  ({ICML})}, ser. Proceedings of Machine Learning Research, vol. 139.\hskip 1em
  plus 0.5em minus 0.4em\relax {PMLR}, 2021, pp. 12\,333--12\,344.

\bibitem{multidmc2025}
F.~Li, H.~Jiang, Z.~Cao, Z.~Liu, Y.~Wang, Z.~Ye, S.~Fan, C.~Li, Y.~Jia, Z.~Qiu,
  M.~Sun, Y.~Wei, and S.~Liu, ``Multi-{DMC}: Deep {Monte-Carlo} with
  multi-stage learning in the card game {UNO},'' in \emph{Proceedings of the
  {IEEE} Conference on Games ({CoG})}, 2025.

\bibitem{vinyals2019alphastar}
O.~Vinyals, I.~Babuschkin, W.~M. Czarnecki, M.~Mathieu, A.~Dudzik, J.~Chung,
  D.~H. Choi, R.~Powell, T.~Ewalds, P.~Georgiev, J.~Oh, D.~Horgan, M.~Kroiss,
  I.~Danihelka, A.~Huang, L.~Sifre, T.~Cai, J.~P. Agapiou, M.~Jaderberg, A.~S.
  Vezhnevets, R.~Leblond, T.~Pohlen, V.~Dalibard, D.~Budden, Y.~Sulsky,
  J.~Molloy, T.~L. Paine, C.~Gulcehre, Z.~Wang, T.~Pfaff, Y.~Wu, R.~Ring,
  D.~Yogatama, D.~W{\"u}nsch, K.~McKinney, O.~Smith, T.~Schaul, T.~Lillicrap,
  K.~Kavukcuoglu, D.~Hassabis, C.~Apps, and D.~Silver, ``Grandmaster level in
  {StarCraft II} using multi-agent reinforcement learning,'' \emph{Nature},
  vol. 575, pp. 350--354, 2019.

\bibitem{ng1999shaping}
A.~Y. Ng, D.~Harada, and S.~J. Russell, ``Policy invariance under reward
  transformations: Theory and application to reward shaping,'' in
  \emph{Proceedings of the 16th International Conference on Machine Learning
  ({ICML})}.\hskip 1em plus 0.5em minus 0.4em\relax Morgan Kaufmann, 1999, pp.
  278--287.

\bibitem{terry2021pettingzoo}
J.~K. Terry, B.~Black, N.~Grammel, M.~Jayakumar, A.~Hari, R.~Sullivan, L.~S.
  Santos, C.~Dieffendahl, C.~Horsch, R.~Perez-Vicente, N.~Williams, Y.~Lokesh,
  and P.~Ravi, ``{PettingZoo}: Gym for multi-agent reinforcement learning,'' in
  \emph{Advances in Neural Information Processing Systems ({NeurIPS})},
  vol.~34, 2021, pp. 15\,032--15\,043.

\bibitem{schulman2017ppo}
\BIBentryALTinterwordspacing
J.~Schulman, F.~Wolski, P.~Dhariwal, A.~Radford, and O.~Klimov, ``Proximal
  policy optimization algorithms,'' 2017. [Online]. Available:
  \url{https://arxiv.org/abs/1707.06347}
\BIBentrySTDinterwordspacing

\bibitem{schulman2016gae}
J.~Schulman, P.~Moritz, S.~Levine, M.~I. Jordan, and P.~Abbeel,
  ``High-dimensional continuous control using generalized advantage
  estimation,'' in \emph{Proceedings of the 4th International Conference on
  Learning Representations ({ICLR})}, 2016.

\end{thebibliography}

\end{document}